\documentclass[journal]{IEEEtran}

\usepackage{cite}
\usepackage{graphicx}
\usepackage{subcaption}
\usepackage{booktabs}
\usepackage{amsmath, amssymb}
\usepackage{makecell}
\usepackage{multirow}
\usepackage{threeparttable}
\usepackage{tikz}
\newcommand*{\circled}[1]{\lower.7ex\hbox{\tikz\draw (0pt, 0pt)%
    circle (.5em) node {\makebox[1em][c]{\small #1}};}}

\ifCLASSINFOpdf
\else
\fi

\begin{document}

\title{Assist-as-needed Hip Exoskeleton Control for Gait Asymmetry Correction via Human-in-the-loop Optimization}

\author{Yuepeng Qian,
        Jingfeng Xiong,
        Haoyong Yu$^{\ast}$,
        and Chenglong Fu$^{\ast}$
        

\thanks{$^{\ast}$ Corresponding authors.}%
\thanks{Y. Qian is with the Department of Biomedical Engineering, National University of Singapore, Singapore 117583, and also with the Department of Mechanical and Energy Engineering, Southern University of Science and Technology, Shenzhen, 518055, China (email: yuepeng.qian@u.nus.edu).}%
\thanks{J. Xiong and C. Fu are with the Department of Mechanical and Energy Engineering, Southern University of Science and Technology, Shenzhen, 518055, China (email: xiongjf2021@mail.sustech.edu.cn, fucl@sustech.edu.cn).}
\thanks{H. Yu is with the Department of Biomedical Engineering, National University of Singapore, Singapore 117583 (email: bieyhy@nus.edu.sg).}
}


\maketitle

\begin{abstract}
Gait asymmetry is a significant clinical characteristic of hemiplegic gait that most stroke survivors suffer, leading to limited mobility and long-term negative impacts on their quality of life. Although a variety of exoskeleton controls have been developed for robot-assisted gait rehabilitation, little attention has been paid to correcting the gait asymmetry of stroke patients following the assist-as-need (AAN) principle, and it is still challenging to properly share control between the exoskeleton and stroke patients with partial motor control. In view of this, this article proposes an AAN hip exoskeleton control with human-in-the-loop optimization to correct gait asymmetry in stroke patients. To realize the AAN concept, an objective function was designed for real-time evaluation of the subject's gait performance and active participation, which considers the variability of natural human movement and guides the online tuning of control parameters on a subject-specific basis. In this way, patients were stimulated to contribute as much as possible to movement, thus maximizing the efficiency and outcomes of post-stroke gait rehabilitation. Finally, an experimental study was conducted to verify the feasibility and effectiveness of the proposed AAN control on healthy subjects with artificial gait impairment. For the first time, the common hypothesis that AAN controls can improve human active participation was validated from the biomechanics viewpoint.
\end{abstract}


\IEEEpeerreviewmaketitle

\section{Introduction}

\IEEEPARstart{S}{troke} is a leading cause of disability worldwide, and its incidence is still increasing with the population aged \cite{feigin2014global,katan2018global}. As a significant clinical characteristic of the hemiplegic gait most stroke survivors suffer, gait asymmetry usually manifests as differences in gait kinematics and dynamics between the healthy and impaired limbs. Long-term gait asymmetry may lead to a series of negative effects on stroke patients, such as reduced walking speed, poor walking stability, increased energy expenditure, increased burden and risk of musculoskeletal injury to the healthy limb because of compensatory movements, all of which result in a significant decrease in their quality of life \cite{patterson2012gait, beyaert2015gait, yen2015using}. Specifically, the root cause of asymmetric gait lies in a wide range of joint movement deficits occurring on the impaired side, such as hip flexion deficit and ankle dorsiflexion deficit \cite{moore1993swing, moseley1993stance}. Then the compensatory movements performed by the healthy limb may further exacerbate the asymmetric gait. Therefore, it is crucial to correct gait asymmetry and restore a normal gait pattern in post-stroke rehabilitation. 

\begin{figure}[tb]
  \centering
  \includegraphics[width=0.70\linewidth]{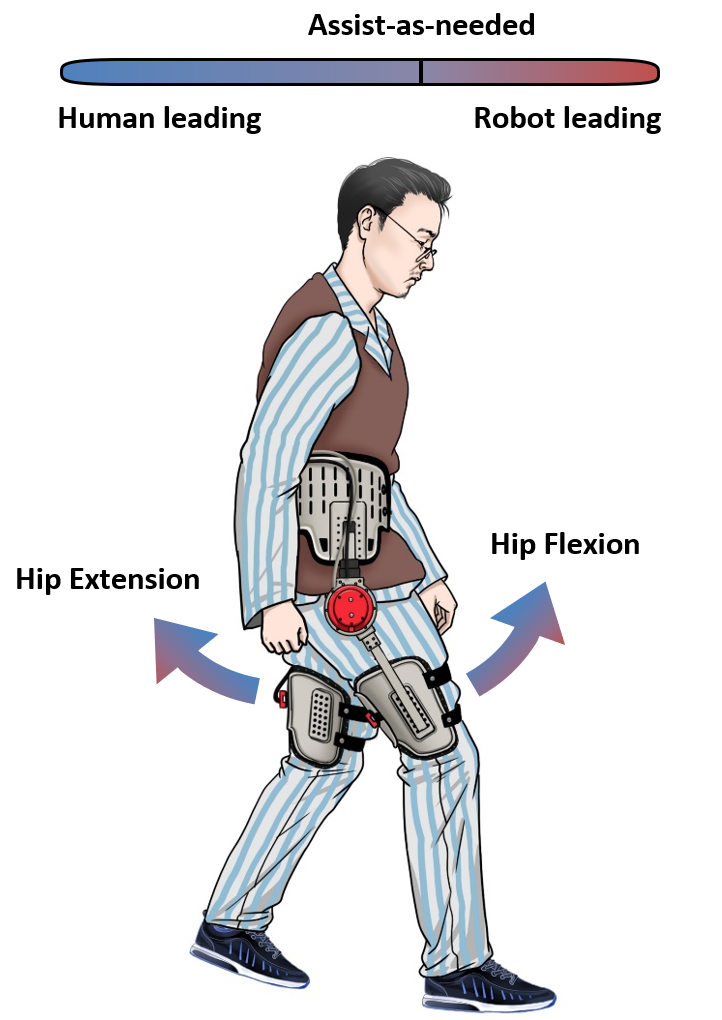}
  \caption{Illustration of patient-robot shared control and assist-as-needed concept during gait rehabilitation with exoskeletons.}
  \label{shared control concept}
\end{figure}

Compared to traditional physiotherapist-assisted rehabilitation, robot-assisted rehabilitation has attracted great attention in recent years because it is capable of providing intensive, precise, and cost-efficient gait training \cite{banala2010novel}. As illustrated in Fig. \ref{shared control concept}, lower-limb exoskeletons cooperate with patients to fulfill the walking task during robot-assisted gait rehabilitation. For patients with severe gait impairment, exoskeletons like ReWalk \cite{esquenazi2012rewalk} and eLEGS \cite{strausser2011development} usually take a leading role in the motion control of lower limbs and guide the patients to walk following a normative gait trajectory with little or no active participation from the patients. However, this assistive control strategy is less suitable for stroke patients who still maintain partial control of lower limbs and may hinder their motor recovery \cite{cai2006implications, israel2006metabolic, duschau2008adaptive}. To maximize the efficiency and outcomes of post-stroke gait rehabilitation, assist-as-needed (AAN) control strategy that allows patients to walk freely and only delivers necessary assistance is desired, which involves consideration for real-time patient-robot cooperation and encourages the active participation of patients. 

To date, how to appropriately share control between the exoskeleton and patients is still a key challenge for robot-assisted gait rehabilitation of stroke patients \cite{diaz2022human, beckerle2017human, li2015continuous}. Most of existing AAN controls were developed based on the impedance concept, such as force field-based control \cite{banala2008robot, banala2010novel, srivastava2014assist} and path control \cite{duschau2009path}. In these control methods, a virtual motion tunnel with impedance walls was constructed along the desired joint trajectory, in which flexible movements under the patient's own control were allowed. However, these control methods obeying the AAN principle still have some limitations. Firstly, the predefined joint trajectory neglects individual gait differences and cannot continuously adapt to changes associated with motor learning during gait rehabilitation. Moreover, the exoskeleton assistance is modulated only in accordance with the trajectory tracking errors, and thus the patient's active participation or intent during real-time interaction is not sufficiently evaluated. Finally, the predefined impedance parameters might only suit specific patients with similar physical conditions, such as muscle strength and body weight. Manual tuning of key control parameters based on observations of the patient's gait performance not only lacks precision, but also needs intensive time and effort. Similar limitations also exist in other error-based adaptive gait rehabilitation controls with AAN properties \cite{hussain2016assist, aguirre2019phase}.

In view of this, velocity field-based control \cite{martinez2018velocity, asl2020field} was proposed for lower-limb or upper-limb rehabilitation, which considered not only the trajectory tracking errors, but also the motion velocity/trend. If the velocity indicated that the leg was converging toward the desired joint trajectory, no exoskeleton assistance would be applied to correct the motion, leading to less interference to patients compared to the force field-based control and path control. However, the velocity field-based control still inherits the drawback of adopting a predefined joint trajectory. To tickle this drawback, the joint trajectory of the healthy limb was adopted as the reference joint trajectory in \cite{zanotto2014adaptive, kawamoto2015modification}, but this approach is also defective considering the compensatory movements frequently occurred on the healthy side. In \cite{jezernik2004automatic, riener2005patient}, the human-exoskeleton interaction force during each gait cycle was measured and evaluated for adapting the reference trajectory of the next gait cycle. In this way, the moving intention of patients was clearly detected and active engagement would be promoted with proper gait leading. Although existing AAN controls have achieved impressive progress in promoting the recovery of patients' walking ability, main efforts still focus on achieving functional walking while little attention has been paid to correcting the gait asymmetry of stroke patients. Besides, the common drawback of utilizing predefined key control parameters is yet well addressed, which is crucial for appropriately sharing control between the exoskeleton and patients. The exoskeleton might take over the walking task or provide insufficient assistance with an improper parameter selection of ANN controls.

In this study, inspired by recent huge success in automatic parameter tuning of wearable robots with human-in-the-loop \cite{zhang2017human, ding2018human, kantharaju2022reducing, park2025stairclimbing, wen2019online, li2021toward, zhang2022imposing}, an AAN hip exoskeleton control with human-in-the-loop parameter tuning is proposed for correcting the gait asymmetry of stroke patients. Different from existing human-in-the-loop controls that only focus on a single objective, such as reducing energy expenditure \cite{zhang2017human, ding2018human, kantharaju2022reducing, park2025stairclimbing} or achieving desired joint trajectory of a robotic prosthesis \cite{gordon2022human, wen2019online, li2021toward, zhang2022imposing}, the objective of the proposed AAN control includes two parts: 1) correct the gait asymmetry (including both the spatial and temporal asymmetry), and 2) stimulate the maximum participation of patients through appropriate shared control. As the first step towards our objective, an adaptive oscillator (AO)-based approach is adopted for continuous gait phase extraction and real-time gait asymmetry detection, which can seamlessly adapt to different levels of gait impairment and changes in gait pace \cite{qian2022adaptive, yan2017oscillator}. Then a P-type iterative learning control is designed to dynamically modulate the exoskeleton assistance for each gait cycle according to the detected spatial and temporal symmetry errors at peak hip flexion. In this way, the assistive control scheme for gait asymmetry correction does not require a reference joint trajectory and gains preliminary AAN property. Furthermore, given unavoidable individual differences in physical conditions and the complexity of human-exoskeleton interaction, human-in-the-loop Bayesian optimization is utilized to identify the key control parameters (that is, the proportional gains in the P-type iterative learning control) on a subject-specific basis. The continuous assessment of the patient's gait performance and active participation allows the exoskeleton assistance to be more precisely tailored and further promotes the active participation of patients.

The major contributions of this work are as follows:
\begin{itemize}
    \item [1)] A novel AAN control scheme for the hip exoskeleton is proposed to correct gait asymmetry in stroke patients, which can adapt to different levels of gait impairment and handle human adaptation through efficient human-in-the-loop parameter tuning.
    \item [2)] Following the AAN principle, an objective function is designed for real-time evaluation of patients' gait performance and active participation, which considers the variability of natural human movement and guides the parameter selection to share control between the exoskeleton and patients appropriately.
    \item [3)] Finally, the feasibility and effectiveness of the proposed method are experimentally validated with human subjects. For the first time, the common hypothesis that AAN controls can improve human active participation is clearly validated from the biomechanics viewpoint.
\end{itemize}

The rest of this paper is organized as follows. Section II introduces the assistive control scheme for gait asymmetry correction and Section III presents the AAN control via human-in-the-loop optimization. In Section IV, the experimental implementation and results are described in detail. Finally, Section V discusses the results and concludes this paper with future works.

\begin{figure*}[tb]
  \centering
  \includegraphics[width=0.95\textwidth]{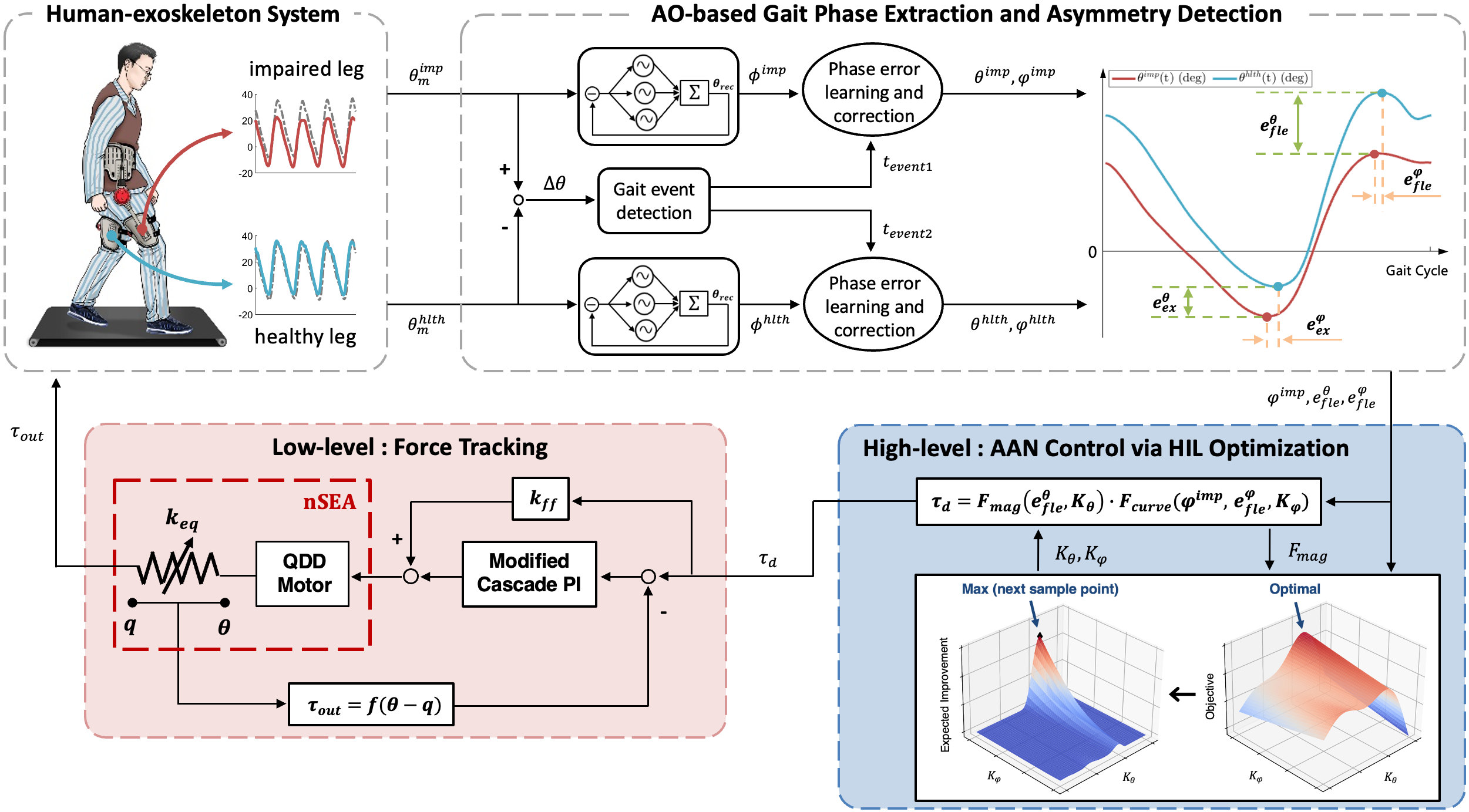}
  \caption{Assist-as-needed hip exoskeleton control for gait asymmetry correction, mainly consisting of the AO-based gait phase extraction and asymmetry detection, the high-level control for assisting as needed via human-in-the-loop optimization, and the low-level control for force tracking.}
  \label{AAN_control}
\end{figure*}

\section{Assistive Control Scheme for Gait Asymmetry Correction}

\subsection{Adaptive Gait Asymmetry Detection}

\subsubsection{Gait Phase Estimation}

As illustrated in Fig. \ref{AAN_control}, the hip angle in the sagittal plane is utilized to extract continuous gait phase in real time with an AO-driven dynamical system. The input of the AO-driven dynamical system is the instantaneous difference between the hip angle measured by the thigh-mounted IMU and the hip angle reconstructed by a Fourier decomposition with finite terms:
\begin{equation}
    e(t) = \theta_{m}(t)- \theta_{rec}(t).
\end{equation}
And the AO-driven dynamical system is given by
\begin{align}
    \Dot{\rho} &= \omega - \psi e(t) \textup{sin}(\rho) \\
    \Dot{\omega} &=- \psi e(t) \textup{sin}(\phi) \\
    \Dot{\alpha}_{j} &= \epsilon \textup{cos}(j \rho)e(t) \quad  (j = 0,...,N_{f}) 
    \label{alpha}\\
    \Dot{\beta}_{j} &= \epsilon \textup{sin}(j \rho)e(t) \quad  (j = 0,...,N_{f}) 
    \label{beta}\\
    \theta_{rec} &= \sum_{j=0}^{N_{f}} \alpha_{j}\textup{cos}(j \rho) + \beta_{j}\textup{sin}(j \rho)\\
    \phi &=\frac{\rm mod(\rho, 2\pi)}{2\pi}
\end{align}
in which $\rho, \omega$ denote the phase and intrinsic frequency of the oscillator, respectively; $\psi, \epsilon$ denote constant coefficients determining the learning speeds. The Fourier decomposition comprises $N_{f}$ pairs of coefficients $\alpha_{j}$ and $\beta_{j}$, which are updated online utilizing (\ref{alpha}) and (\ref{beta}). Finally, the gait phase estimation $\phi$ can be generated by normalizing the oscillator phase $\rho$ into the interval [0,1), corresponding to 0-100$\%$ of the gait cycle.

\subsubsection{Gait Event Detection and Phase Synchronization}

The gait phase $\phi$ estimated by the AO-driven dynamical system can not be used in exoskeleton controls directly due to inevitable mismatches between the estimated gait phase and the progress of real gait cycles. For accurate synchronization of the estimated gait phase to the user's gait, the gait phase at a landmark gait event is expected to be locked to a specific constant in each stride. Different from existing hip exoskeleton controls \cite{kim2018autonomous, zhang2019admittance} that adopted the peak hip flexion/extension (PHF/PHE) as the landmark gait event, the hip angle difference $\triangle\theta(t)$ is utilized to detect the landmark gait event for the convenience of analyzing the temporal asymmetry at PHF/PHE:
\begin{equation}
    \triangle \theta(t) = \theta^{imp}_{m}(t) - \theta^{hlth}_{m}(t)
\end{equation}
in which $\theta^{imp}_{m}(t)$, $\theta^{hlth}_{m}(t)$ denote the IMU-measured hip angles of the impaired and healthy sides. The maximum and minimum of $\triangle\theta$ are defined as the landmark gait event for the impaired side and the healthy side, respectively. When the landmark gait events are detected, the gait phase of the healthy side directly estimated by the AO-driven dynamical system is recorded:
\begin{align}
\phi^{hlth}(\triangle\theta_{max}) & = \phi^{hlth}(t_{k1}) \\
\phi^{hlth}(\triangle\theta_{min}) & = \phi^{hlth}(t_{k2})
\end{align}
where $t_{k1}, t_{k2}$ represent the timings when $\triangle\theta_{max}$ and $\triangle\theta_{min}$ are detected, respectively. 

According to the definition of temporal symmetry, the two landmark gait events ($\triangle\theta_{max}, \triangle\theta_{min}$) should occur half a gait cycle apart. In other words, the gait phase of the healthy side should be 0.5 when the landmark gait event of the impaired side ($\triangle\theta_{max}$) is detected. Thus, the synchronization errors of the gait phase can be obtained by:
\begin{align}
    e^{imp}_{\phi}(t_{k1}) &= \phi^{hlth}(t_{k1}) - 0.5 \\
    e^{hlth}_{\phi}(t_{k2}) &= -\phi^{hlth}(t_{k2})
\end{align}
and then corrected through the following equations \cite{yan2017oscillator,qian2022adaptive}:
\begin{equation}
    \Dot{\delta}_{\phi} = K_{\phi} \cdot[e_{\phi}(t_{k})-\delta_{\phi}] \cdot e^{-\omega(t)(t-t_{k})}
\end{equation}
\begin{equation}
    \varphi(t) = \mathrm{mod}(\phi(t)-\delta_{\phi}(t), 1)
\end{equation}
where $K_{\phi}$ denotes an adjustable coefficient, $t_{k}$ denotes the timing when the last landmark gait event is detected, $\omega$ represents the frequency tracked by the AO, and the corrected gait phase estimation is noted as $\varphi(t)$. After the correction of phase synchronization errors, the estimated gait phase of the healthy side is accurately synchronized to real gait cycles while the estimated gait phase of the impaired side advances or lags half a gait cycle compared to the healthy side, paving the way for quantifying temporal gait asymmetry and providing suitable walking assistance.

\subsubsection{Gait Asymmetry Analysis}

Gait symmetry includes spatial symmetry and temporal symmetry. As shown in Fig. \ref{AAN_control}, the former requires geometric values at the same landmark gait event to match while the latter requires the landmark gait event of both sides to occur at the same gait phase. In each gait cycle, the peak hip angle and corresponding gait phase $[\theta_{p}(i), \varphi_{p}(i)]$ ($i$ = 1, 2, ...) are recorded for the analysis of gait symmetry. To attenuate the influence of normal gait variations, the averages of peak hip flexion/extension angle and corresponding gait phase for the last $N_s$ strides are utilized in the hip exoskeleton control. Thus, the estimation of $\theta_{p}(i), \varphi_{p}(i)$ for the $i$-th stride can be calculated by
\begin{equation}
    \theta_{p,est}(i)=\frac{1}{N_s}\sum_{n=0}^{N_s-1}\theta_{p}(i-n)
\end{equation}
\begin{equation}
    \varphi_{p,est}(i)=\frac{1}{N_s}\sum_{n=0}^{N_s-1}\varphi_{p}(i-n)
\end{equation}
and then the spatial and temporal symmetry errors between the healthy and impaired sides at PHF/PHE can be yielded by
\begin{equation}
    e^{\theta}_{p}=\theta_{p,est}^{hlth}(i)-\theta_{p,est}^{imp}(i)
\end{equation}
\begin{equation}
    e^{\varphi}_{p}=\varphi_{p,est}^{hlth}(i)-\varphi_{p,est}^{imp}(i)
\end{equation}
in which $p$ denotes peak hip flexion/extension (PHF/PHE).

\subsection{Assistive Control for Gait Asymmetry Correction}

For stroke patients with hemiplegic gait, the leading cause of gait asymmetry at hip joints is the hip flexion deficit that occurs on the impaired side, while the hip extension deficit usually occurs on the healthy side due to abnormal intrajoint coupling. In this context, the hip exoskeleton control presented herein mainly focuses on correcting the hip flexion deficit that occurred on the impaired side and then restoring the gait symmetry of stroke patients during the whole gait cycle. The assistive torque within each gait cycle is formulated by the following equation
\begin{equation}
    \tau_d (\varphi^{imp}, e^{\theta}_{fle}, e^{\varphi}_{fle}) = F_{mag}(e^{\theta}_{fle}) \cdot F_{curve}(\varphi^{imp},e^{\varphi}_{fle})
\end{equation}
where the detected asymmetric errors, $e_{fle}^{\theta}, e_{fle}^{\varphi}$, are utilized to determine the magnitude and peak timing of assistive torque through P-type iterative learning control. Then the desired assistive torques would be generated by a nonlinear series elastic actuator (nSEA)-driven hip exoskeleton with high accuracy and robustness.

\subsubsection{Computing the magnitude of assistive torque}

To compensate hip flexion deficit that occurred on the impaired side ($e^{\theta}_{fle}\geq0$), the magnitude of assistive torque is determined according to the spatial asymmetry between the impaired and healthy legs:
\begin{equation}
    F_{mag}(i)=\lambda_{\theta} \cdot F_{mag}(i-1) + (1-\lambda_{\theta}) \cdot K_{\theta} \cdot e^{\theta}_{fle}
\label{F_fle}    
\end{equation}
in which $\lambda_{\theta} < 1$ denotes a forgetting factor, and $K_{\theta}$ denotes as a proportion gain for the angle difference. Specially, no exoskeleton assistance would be applied to the impaired leg if $e^{\theta}_{fle}<0$.

\begin{figure}[tb]
  \centering
  \includegraphics[width=0.42\textwidth]{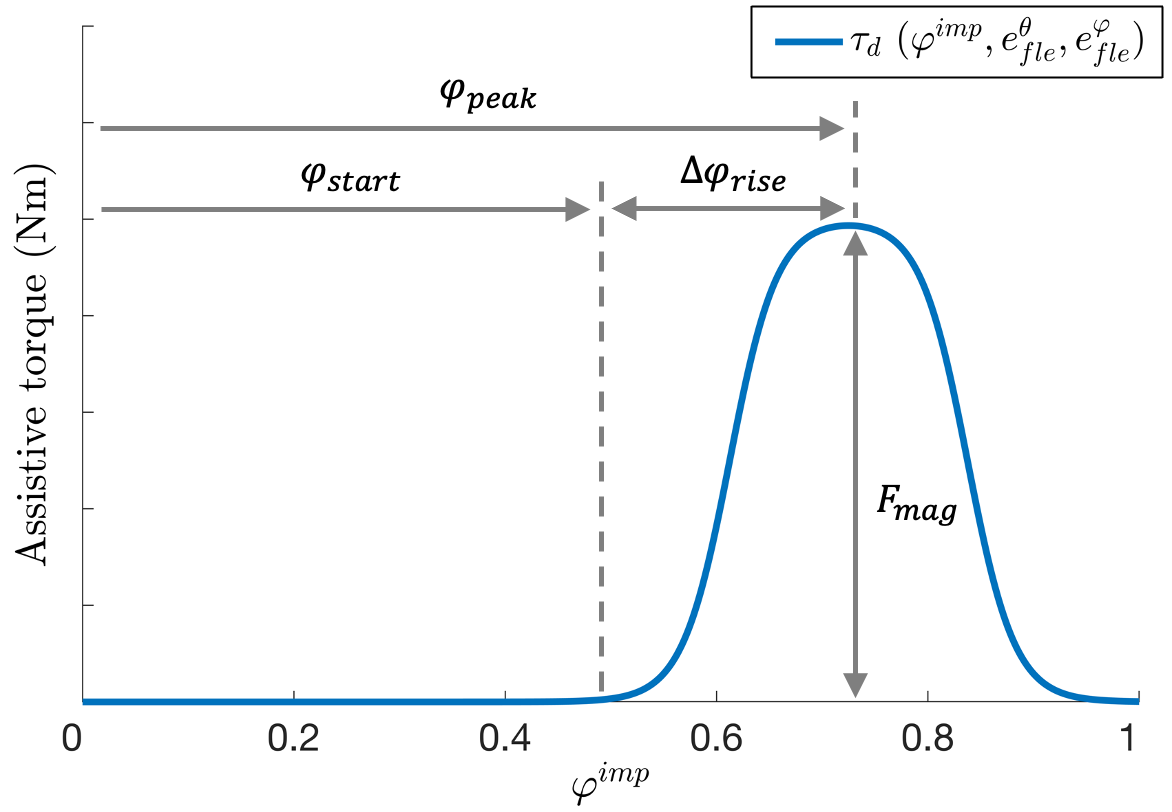}
  \caption{Assistive torque $t_{d}$ as a function of the gait phase $\varphi^{imp}$.}
  \label{force_profile}
\end{figure}

\subsubsection{Generating the curve of assistive torque}

As illustrated in Fig. \ref{force_profile}, the reference torque curve is generated based on a parameterized unimodal curve $h(\varphi, \triangle\varphi_{start})$. Specifically, the rising time $\triangle \varphi_{rise}$ is predetermined, and thus the peak timing of assistive torque $\varphi_{peak}$ is indirectly adjusted by choosing different starting phase value $\triangle \varphi_{start}$:
\begin{equation}
    \begin{aligned}
    h(\varphi, \varphi_{start}) =\frac{1}{2} \Bigg\{ \tanh \bigg\lbrack a \left({\kappa(\varphi, \varphi_{start}) - \frac{1}{2}}\right)\bigg\rbrack  \\
    +\tanh \bigg\lbrack a \left({\kappa(\varphi, \varphi_{start})-\frac{3}{2}}\right)\bigg\rbrack \Bigg\}
    \end{aligned}
\end{equation}
in which $a>0$ and
\begin{equation}
    \kappa(\varphi, \varphi_{start}) =\frac{\varphi-\varphi_{start}}{\triangle\varphi_{rise}}.
\end{equation}

To correct the temporal asymmetry between the impaired and healthy legs, the peak timing of assistive torque is dynamically adjusted according to the phase difference at peak hip flexion:
\begin{equation}
   \varphi_{start}(i) = \lambda_{\varphi} \cdot \varphi_{start}(i-1) + (1-\lambda_{\varphi})\cdot ( \varphi_{start}^{init} -K_{\varphi}\cdot e^{\varphi}_{fle})
\end{equation}
in which $\lambda_{\varphi}<1$ denotes a forgetting factor, $\varphi_{start}^{init}$ is the initial starting phase value predefined according to the normal human gait, and $K_{\varphi}$ denotes a proportion gain for the phase difference. Considering the patients' safety during assistive walking, the adjusted starting phase $\varphi_{start}$ is restricted to the interval $\big\lbrack \varphi_{start}^{init}-0.05,\ \varphi_{start}^{init}+0.05 \big\rbrack$. Then the assistive torque curve is generated by 
\begin{equation}
    F_{curve}(\varphi^{imp},e^{\varphi}_{fle}) =  h(\varphi^{imp}, \varphi_{start}).
\end{equation}

\subsubsection{Torque control of nSEA-driven hip exoskeleton}

In this study, a unilateral nSEA-driven hip exoskeleton is adopted to assist the impaired limb of stroke patients, which provides assistive torques of up to 19.8 Nm and features excellent transparency/back-drivability \cite{qian2022toward}. Specifically, the output torque of the nSEA-driven hip exoskeleton is controlled basically by regulating the deflection of the rotary series elastic element and the angle of deflection can be calculated by:
\begin{equation}\label{deflection_angle}
\alpha=\theta - q
\end{equation}
in which ${\theta}, {q}$ denote the rotation angle of the input and output sides of the rotary series elastic element.

Then a modified cascade PI controller is utilized to carry out assistive torques with high accuracy and robustness, and the control law can be represented as

\begin{equation} \label{Eq_controller}
    {\tau _m} = \left( {{k_{vp}} + {k_{vi}}\frac{1}{s}} \right)\left[ {\left( {{k_{pp}} + {k_{pi}}\frac{1}{s}} \right)\left( {{\alpha _d} - \alpha } \right) - \dot \theta} \right]+ {k_{f}}\cdot{\tau_d}.
\end{equation}
in which $\tau_m$ denotes the motor torque, $k_{vp}, k_{vi}, k_{pp}, k_{pi}$ denote the gains for the position-loop and velocity-loop PI controllers, $k_{f}$ denotes the gain for the feedforward term, and $s$ denotes the Laplace operator. Its effectiveness has been experimentally validated with human subjects, and more details about the nSEA-driven Hip exoskeleton and its torque control can be found in our previous publications \cite{qian2022toward,qian2022mechatronics}.

\section{Assist-as-needed Control via Human-in-the-loop Optimization}

\subsection{Problem Formulation}
Unlike existing human-in-the-loop controls of wearable robots that focus on reducing the user's energy expenditure \cite{zhang2017human, ding2018human, kantharaju2022reducing, gordon2022human} or achieving the desired joint trajectory of prosthesis \cite{wen2019online, li2021toward, zhang2022imposing}, the preliminary objective of the proposed hip exoskeleton control is to correct the gait asymmetry of stroke patients, but this is not the only objective. In keeping with the AAN principle, the proposed control should restore a symmetric gait through as little assistance as possible, thereby encouraging the patients' active participation and helping them learn to overcome the impairment, at least partially.

\begin{figure}[tb]
  \centering
  \includegraphics[width=0.4\textwidth]{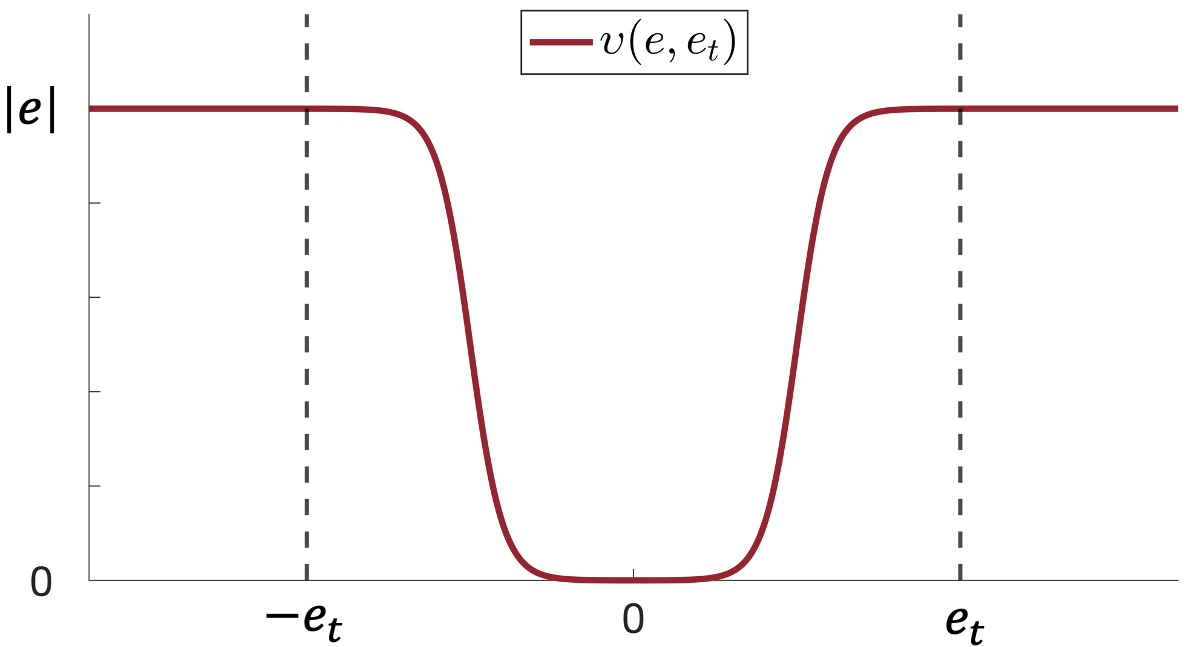}
  \caption{Activation function $\upsilon$ for the symmetry errors, such as $e_{fle}^{\theta}$, $e_{fle}^{\varphi}$.}
  \label{activation_fun}
\end{figure}

As introduced in Section III-B, the magnitude and peaking timing of assistive torque are determined by an error-based control with forgetting factors and the exoskeleton assistance would be reduced when the detected symmetry errors are small, which can be regarded as a primary AAN control for gait asymmetry correction. However, the unavoidable individual differences in physical conditions and the complexity of human-exoskeleton interaction can lead to different responses to the same level of exoskeleton assistance (that is, the same proportion gains, $K_{\theta}, K_{\varphi}$). To provide suitable assistance during gait asymmetry correction, the key control parameters ($K_{\theta}, K_{\varphi}$) should be optimized based on real-time evaluation of the patient's gait performance and active participation, and thus the objective function can be defined as
\begin{align}
    O_{AAN} & = -\gamma \odot E^{T} \\
      & = -\gamma \odot \Big\lbrack \upsilon(e_{fle}^{\theta}, e_{t}^{\theta}), \upsilon(e_{fle}^{\varphi}, e_{t}^{\varphi}), F_{mag}\Big\rbrack^{T}
\end{align}
\label{Objective_AAN}
where $\gamma \in R^{3\times1}$ is a vector of scaling factors that weighs the relative cost of the symmetry errors and exoskeleton assistance (indirectly represents the patient’s active participation), and $\odot$ is the Hadamard product of two vectors. Specially, $\upsilon(\cdot)$ is the activation function for the symmetry errors as illustrated in Fig. \ref{activation_fun}
\begin{equation}
    \upsilon(e,e_{t}) = \vert e\vert \cdot \bigg\lbrack\frac{1}{1+exp(-\frac{20}{e_{t}} \cdot e +10)} + \frac{1}{1+exp(\frac{20}{e_{t}} \cdot e +10)}\bigg\rbrack
\label{activation_equation}
\end{equation}
where $e$ denotes the symmetry error and $e_{t}$ represents the tolerated error considering the natural range of human movement variability. Thus the problem is posed to find the optimal control parameter set $x^{+} = \big\lbrack K_{\theta}, K_{\varphi} \big\rbrack$, let
\begin{equation}
    O_{AAN}(x^{+}) = \max_{x \in A} O_{AAN}(x)
\end{equation}
where $A \subset R^{2}$ is the feasible parameter region.

\subsection{Bayesian Optimization}

Bayesian optimization (BO) was employed to optimize the two-parameter set $x = \big\lbrack K_{\theta}, K_{\varphi} \big\rbrack$ since it is an efficient and noise-tolerant global optimization method, and detailed implementation referred to the tutorials introduced in \cite{brochu2010tutorial, frazier2018tutorial}. As the first step, the BO was initialized by evaluating the optimization objective with five predefined sets of control parameters, which were selected from the feasible parameter region $A$ with evenly spaced intervals to avoid biased sampling that might lead to premature convergence and local optimum [see Fig. \ref{protocol}(b)]. After the initialization period, the landscape of optimization objective $O_{AAN}(x)$ was calculated using the Gaussian process (GP). Then the new parameter set was automatically selected by maximizing the acquisition function, Expectation Improvement (EI). Once the new sample of the optimization objective and its corresponding parameter set were added to the dataset, the optimization objective landscape was updated for selecting the next sampling parameter set with the acquisition function, EI.

In the GP, the posterior distribution of the optimization objective is estimated as a function of the control parameter set $x$. The prior of the GP is characterized using the mean $\mu(x)$ and the covariance matrix $m(x,x')$. As is common practice, the mean is set to be zero and the squared exponential kernel is adopted for the covariance matrix
\begin{equation}
    m(x,x')=\sigma^{2}\cdot \mathrm{exp}\Big(-\frac{1}{2}(x-x')L(x-x')^{T}\Big)
\end{equation}
where $\sigma$ is the optimization objective variance and
\begin{equation}
    L=
    \begin{bmatrix}
    l_{1} & 0 \\ 0 &l_{2}    
    \end{bmatrix}
\end{equation}
 Specifically, $l_{1},l_{2}$ are the length scale parameters of the two control parameters, $K_{\theta}, K_{\varphi}$.

In practice, noise-free observations are rarely possible. Thus it is assumed that the samples of the optimization objective have an additive, independent, and identically distributed noise $N(0, \sigma_{noise}^{2})$, and they can be expressed as
\begin{equation}
    y(x)=O_{AAN}(x)+w, \ w\sim N(0,\sigma_{noise}^{2})
\end{equation}
where $\sigma_{noise}^{2}$ is the noise variance. Based on the prior of the GP and the collected dataset $D=\{X, Y\}$ \big($X=\{x_{1}, \ldots, x_{n}\}; Y=\{y_{1}, \ldots,y_{n}$\}\big), the posterior distribution of the optimization objective at a parameter set $x_{*}$ could be calculated as
\begin{equation}
    P\big(y_{*}|D,x_{*}\big)=N\big(\mu(x_*),\sigma^{2}(x_*) \big)
\end{equation}
\begin{equation}
    \mu(x_{*})=m^T_{n} \cdot (M+ \sigma^{2}_{nosie} \cdot I)^{-1} \cdot 
\end{equation}
\begin{equation}
    \sigma(x_{*})=m(x_{*},x_{*})-m^{T}_{n} \cdot \big \lbrack M + \sigma^{2}_{x_{*}} \big \rbrack^{-1} \cdot m_{n} 
\end{equation}
where $m_{n}$ is the kernel vector and M is the kernel matrix, given by
\begin{equation}
    m_{n}= \big \lbrack m(x_*, x_1),\ldots, m(x_*, x_n) \big \rbrack
\end{equation}
\begin{equation}
    M=
    \begin{bmatrix}
    m(x_1, x_1) & \cdots & m(x_1, x_n) \\ 
    \vdots & \ddots & \vdots \\ 
    m(x_n, x_1) & \cdots & m(x_n, x_n)
    \end{bmatrix}
\end{equation}
All hyperparameters ($\sigma, \sigma_{noise}, l_1, l_2$) were optimized by maximizing the marginal log-likelihood of the collected dataset $D$ at each iteration.

The acquisition function, EI, selects the next parameter set to be evaluated by maximizing the expected improvement using the following equation
\begin{equation}
    EI(x_*)= \big( \mu(x_*)-y^{+}-\zeta \big) \cdot cdf(z) + \sigma(x_*) \cdot pdf(z)
\end{equation}
where 
\begin{equation}
    z = \frac{\mu(x_*)-y^{+}-\zeta}{\sigma(x_*)}
\end{equation}
Specially, both $EI$ and $z$ are set to be zeros if $\sigma(x_*)$ is zero. $y^{+}$ is the best sample of the optimization objective during all previous iterations. $\zeta$ is a non-negative trade-off parameter that balances exploration and exploitation \cite{lizotte2008practical}. $cdf(\cdot)$ and $pdf(\cdot)$ denote the cumulative distribution function and probability distribution function of the normal distribution. Then the next parameter set $x_{next}$ to be evaluated is generated by 
\begin{equation}
    x_{next} = \mathrm{argmax}_{x_*} EI(x_*).
\end{equation}

\subsection{Human Variability and Stopping Criterion}

To reduce the influence of inherent walking variability, the key control parameters ($K_{\theta}, K_{\varphi}$) are tuned every twenty gait cycles (one episode). Besides, only data collected from the last ten gait cycles of each episode is used to evaluate gait performance and human participation, considering that it takes several steps for users to adapt to new control parameters and stabilize their gait when a new episode starts. If the change of the automatically selected control parameters in three consecutive episodes is less than 0.03, the control parameter optimization is considered finished and then the proposed hip exoskeleton control would continuously assist the user with the optimal parameters.

In this study, we also take into account the natural range of human movement variability in the evaluation of gait performance and control parameter optimization, by introducing the tolerated errors ($e_{t}^{\theta}, e_{t}^{\varphi}$) for the spatial and temporal symmetry errors. Specifically, the tolerated errors are determined referring to \cite{winter1984kinematic, kadaba1990measurement}
\begin{equation}
    e_{t}^{\theta} = 2.5^{\circ}, \ e_{t}^{\varphi}= 1.5 \%
\end{equation}
Through the activation function $\upsilon(\cdot)$ defined in (\ref{activation_equation}), the optimization algorithm would disregard the natural human movement variability and focus on encouraging the patient's active participation when the gait performance is satisfactory.

\section{Experiments and Results}

\begin{figure*}[tb]
  \centering
  \includegraphics[width=0.95\textwidth]{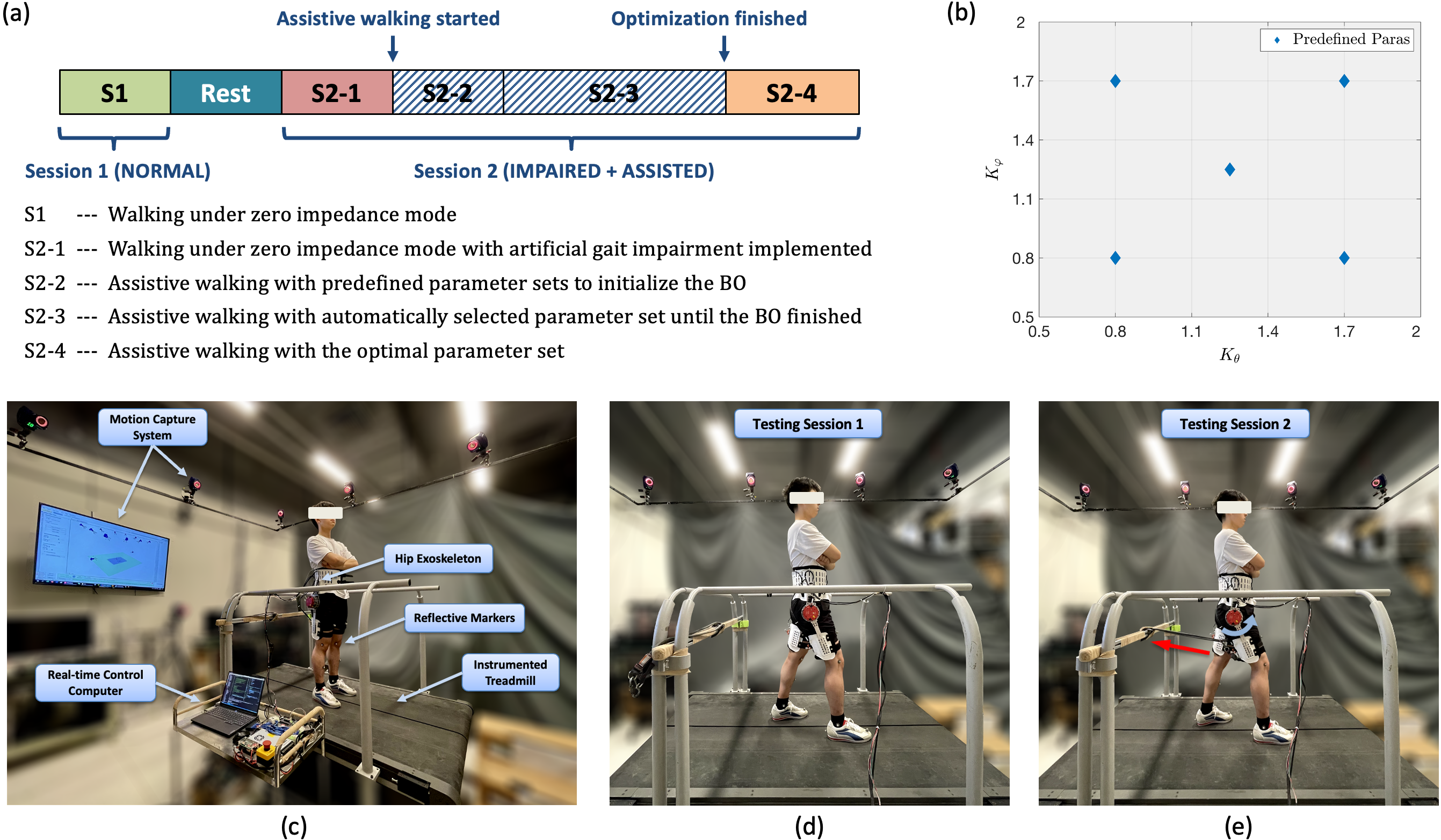}
  \caption{(a) Experimental protocol. (b) Feasible parameter region $A$ and five predefined sets of control parameters for the initialization of BO. (c) Overview of experimental environment and setup. (d)-(e) Experimental scenarios of different testing sessions. Specifically, the height of elastic ropes that are used for creating artificial gait impairment can be adjusted to accommodate the height of subjects. }
  \label{protocol}
\end{figure*}

\begin{figure*}[tb]
  \centering
  \includegraphics[width=0.92\textwidth]{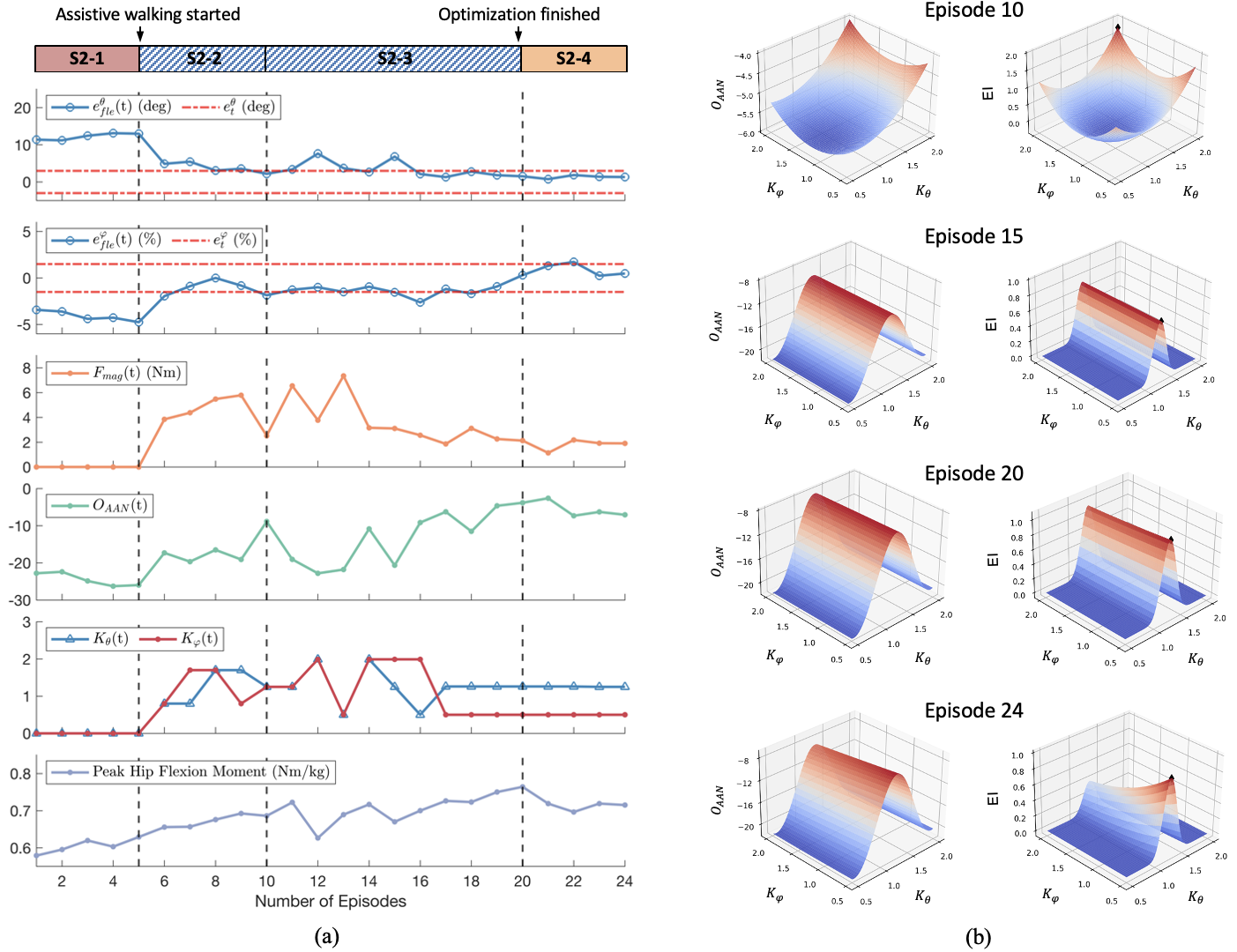}
  \caption{Experimental results over the number of episodes in the testing session 2. (a) Evolution of different human-exoskeleton system states as control parameters were updated. (b) Landscapes of the optimization objective (right) and the expected improvement (left) at different episodes.}
  \label{results_demo}
\end{figure*}

\subsection{Participants and Simulated Hemiplegic Gait}
To validate the efficacy of the proposed AAN hip exoskeleton control in improving gait symmetry and active participation of patients, an experimental study involving five healthy subjects (height 168 $\pm$ 3 cm; weight 60.8 $\pm$ 4.8 kg; age 23.8 $\pm$ 1.6 years; and mean $\pm$ SD) was conducted with approval from the Institutional Review Board at National University of Singapore (NUS-IRB reference: LH-20-012). 

The experimental study involved a simulated hemiplegic gait, and using artificial gait impairments to study robot-aided gait rehabilitation has been widely adopted and tested in \cite{emken2005robot, emken2007human, aguirre2020lower, qian2022adaptive}. To create artificial gait impairment in the healthy subjects, elastic ropes with a stiffness of 2.8 kN/m were attached to the thigh of the simulated impaired leg, providing resistance for hip flexion and inducing a hemiplegic gait pattern as shown in Fig. \ref{protocol}(e). Experimental results suggested that the simulated hemiplegic gait features decreased peak hip flexion on the impaired side, decreased peak hip extension on the healthy side, and alterations in the associated gait phase at peak hip flexion/extension. These gait features well align with the characteristics of hemiplegic gait observed post-stroke, thus validating the feasibility of employing artificial gait impairments in this proof-of-concept study \cite{olney1994temporal}.

\subsection{Experimental Setup}

The experimental environment and setup are illustrated in Fig. \ref{protocol}. Specifically, the proposed hip exoskeleton control was implemented on the Robot Operation System (ROS) with an update rate of 250 Hz. The angle of hip flexion/extension, $\theta_{m}(t)$, was measured by the thigh-mounted IMUs (MTi-630, Xsens, Netherlands) at a frequency of 400 Hz, and fed back to the control system via the controller area network (CAN) bus. Besides, gait kinematic and dynamic data were recorded by a motion capture system with 12 cameras (Raptor-12HS, Motion Analysis, USA) at a frequency of 120 Hz and an instrumented treadmill (Bertec, USA) at a frequency of 1200 Hz. Then biomechanical analysis was performed with the motion analysis software, Visual 3D (C-motion, USA), to provide reliable information for the analysis of subjects' active participation during gait rehabilitation.

\subsection{Experimental Protocol}

As illustrated in Fig. \ref{protocol}(a), each subject was asked to walk on the treadmill for two testing sessions. Given that 0.8 m/s is regarded as the baseline speed for community life and normal human walking \cite{winter2009biomechanics, bohannon2011normal}, the treadmill speed was set to 0.8 m/s for all testing sessions, and a fixed walking speed also helped to exclude the effects of walking speed on the evaluation of gait performance and human participation. Besides, to prevent potential confounding effects of fatigue, a rest of 5-10 minutes was performed between different testing sessions. A more detailed introduction about the two testing sessions is as follows:

\subsubsection{Testing session 1}
 During testing session 1, the subjects walked with the exoskeleton control set to zero impedance mode for five episodes (100 gait cycles), aiming to generate the normal gait information of subjects as a baseline for comparison purposes. The processed experiment results of testing session 1 are noted as NORMAL for brevity in the following sections.
 
\subsubsection{Testing session 2}
As illustrated in Fig. \ref{protocol}(a), testing session 2 could be divided into four sub-sessions (S2-1 to S2-4), and assistive torques from the hip exoskeleton were applied from S2-2 onward. Specifically, the sub-session S2-1 and S2-2 lasted for five episodes while the sub-session S2-3 ended until the online optimization finished (that is, the stopping criterion was satisfied). For ease of comparison, the processed experiment results of S2-1, S2-2, and the last three episodes of S2-3 are noted as IMPAIRED, ASSISTED (Predefined Paras), and ASSISTED (Optimal Paras), respectively, in the following sections. Specifically, during the sub-session S2-2, the five predefined sets of control parameters were explored in random order to avoid possible evaluation bias, and the feasible parameter region $A$ was determined according to our experimental experiences of manually tuning the control parameters for different subjects, as well as considerations for the effectiveness and safety of exoskeleton assistance.

\begin{table*}
\centering
\caption{Evaluation results of the subjects' gait performance and active participation}
\begin{tabular}{cccccccc} 
\toprule
  Experimental Conditions  & SAP$_{fle}$\; \rm{(deg)} & TAP$_{fle}$\; ($\%$) & SAP$_{ex}\ \rm{(deg)}$ & TAP$_{ex}$\; ($\%$) & SI$_{ROM}\ (\%)$ & SI$_{TOA}$ & HPI \\
  \midrule
   NORMAL                      & 0.58 $\pm$ 0.94  & -1.40 $\pm$ 0.52 & 0.92 $\pm$ 0.71 & -0.51 $\pm$ 0.81 & -1.18 $\pm$ 2.94 & 0.985 $\pm$ 0.003 & 1.000 $\pm$ 0.000 \\
   IMPAIRED                    & 12.14 $\pm$ 1.15 & -2.90 $\pm$ 0.81 & 8.44 $\pm$ 1.40 & -0.27 $\pm$ 1.01 & 9.24 $\pm$ 2.21  & 0.754 $\pm$ 0.034 & 1.442 $\pm$ 0.073 \\
    ASSISTED (Predefined Paras) & 3.40 $\pm$ 0.29  & -0.64 $\pm$ 0.40 & 4.74 $\pm$ 1.12 & -3.17 $\pm$ 0.60 & -2.92 $\pm$ 2.77 & 0.926 $\pm$ 0.018 & 1.612 $\pm$ 0.131 \\
   ASSISTED (Optimal Paras)   & 2.58 $\pm$ 0.60  & 0.26 $\pm$ 0.46  & 3.37 $\pm$ 1.09 & -1.35 $\pm$ 1.28 & -1.60 $\pm$ 2.21 & 0.957 $\pm$ 0.011 & 1.774 $\pm$ 0.138 \\
  \bottomrule
  \end{tabular}
  \label{results_table}
\end{table*}



\begin{figure*}[tb]
  \centering
  \includegraphics[width=1\textwidth]{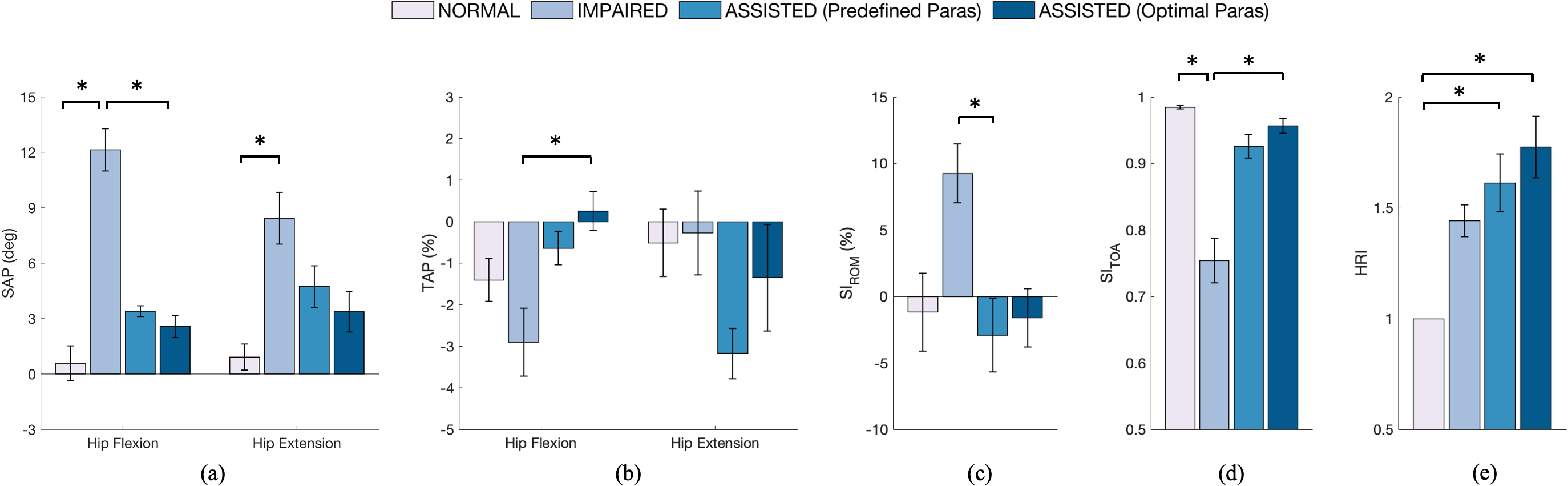}
  \caption{Evaluation results across different experimental conditions and subjects. All results are presented with mean $\pm$ sem and the asterisks indicate statistically significant differences ($p <$ 0.05). (a) Spatial asymmetry at peak hip flexion/extension (SAP). (b) Temporal asymmetry at peak hip flexion/extension (TAP). (c) Symmetry index for the range of joint motion (SI$_{ROM}$). (d) Symmetry index for the time trajectory of joint angle (SI$_{TOA}$). (e) Human participation index (HPI).}
  \label{results_bar}
\end{figure*}

\subsection{Evaluation Metrics}
To quantify the gait performance and active participation of subjects, five different metrics were utilized:

\subsubsection{Spatial asymmetry at peak hip flexion/extension (SAP)}
\begin{equation}
    SAP = e_{p}^{\theta}= (\theta_{p,est}^{hlth}-\theta_{p,est}^{imp})
\end{equation}

\subsubsection{Temporal asymmetry at peak hip flexion/extension (TAP)}
\begin{equation}
    TAP = 100 \cdot e_{p}^{\varphi}=100 \cdot (\varphi_{p,est}^{hlth}-\varphi_{p,est}^{imp})
\end{equation}

\subsubsection{Symmetry index for the range of joint motion ($SI_{ROM}$)} 
\begin{equation}
    ROM = \theta_{p,fle}-\theta_{p,ex}
\end{equation}
\begin{equation}
    SI_{ROM} = \frac{ROM^{hlth}-ROM^{imp}}{avg(ROM^{hlth}, ROM^{imp})}\cdot 100
\end{equation}

\subsubsection{Symmetry index for the time trajectory of joint angle ($SI_{TOA}$)}
Despite the three discrete variables defined above, we also introduced a continuous variable, $SI_{TOA}$, to reflect the gait symmetry during the whole gait cycle or continuous evaluation period, which refers to the "variance accounted for" (VAF) metric \cite{van2004identification} and is defined as
\begin{equation}
    SI_{TOA} = 1 - \frac{var(\theta^{hlth}-\theta^{imp})}{var(\theta^{hlth})}
\end{equation}
where the time trajectory generated from the healthy side, $\theta^{hlth}$, is adopted as the reference, and both time trajectories are indexed to the estimated gait phase, i.e., $\theta^{imp} = \theta^{imp}(\varphi^{imp})$ and $\theta^{hlth} = \theta^{hlth}(\varphi^{hlth})$. $SI_{TOA}$ = 1 indicates that perfect symmetry is achieved.

\subsubsection{Human participation index (HPI)}
In addition to the gait symmetry indexes, the peak hip flexion moment of the impaired side in each gait cycle was measured with the motion capture system and instrumented treadmill, as the ground truth of human participation. However, due to large variations in the subject's physical conditions, there were obvious differences in absolute value or even normalized value of the hip moment among different subjects. In this context, to clearly demonstrate changes in human participation, the average of peak hip flexion moment normalized by body weight during the testing session 1, $\tau_{hip, NORMAL}$, is adopted as the reference and thus the relative level of human participation can be quantified by
\begin{equation}
    HPI = \frac{\tau_{hip}}{\tau_{hip,NORMAL}}
\end{equation}
where $\tau_{hip}$ denotes the average of peak hip flexion moment normalized by body weight during different experimental sessions (NORMAL, IMPAIRED, and ASSISTED with predefined or optimal paras).

\subsection{Statistical Analysis}

To compare the changes in subjects' gait performance and active participation across different experimental conditions, the Kruskal-Wallis test was utilized due to the relatively small number of subjects (n = 5). If a statistically significant difference was detected among different experimental conditions (defined as $p < 0.05$), post-hoc Tukey’s honest significant difference (HSD) tests were conducted to identify specific pairwise differences. All statistical analysis was performed using MATLAB 2023a.

\subsection{Results}

As can be seen from Fig. \ref{results_demo}, the gait asymmetry at peak hip flexion was gradually corrected and the subject's active participation was also promoted as the key control parameters were updated. The online optimization could be finished in an average of 312 $\pm$ 120 gait cycles (or 5.9 $\pm$ 3.2 minutes), which equals the duration of S2-2 and S2-3 shown in Fig. \ref{results_demo}(a), and no more updates of the key control parameters were needed during stable walking (S2-4). Detailed evaluation results of the subjects' gait performance and active participation are summarized in Table \ref{results_table} and illustrated in Fig. \ref{results_bar}. With effective assistance from the hip exoskeleton, all evaluation metrics of gait symmetry (SAP, TAP, SI$_{ROM}$, and SI$_{TOA}$) were obviously improved compared to IMPAIRED. Besides, it is worth noting that ASSISTED (Optimal Paras) outperformed ASSISTED (Predefined Paras) in terms of all evaluation metrics of gait symmetry and human participation, suggesting the effectiveness and necessity of optimizing the key control parameters on a subject-specific basis.

Although exoskeleton assistance was only provided to compensate for the hip flexion deficit that occurred on the impaired side, experimental results (SAP, TAP) suggested that gait symmetry was effectively improved at both the peak of hip flexion and extension. This is because the leading cause of gait asymmetry at hip joints is the hip flexion deficit that occurs on the impaired side, while the hip extension deficit usually occurs on the healthy side due to abnormal intrajoint coupling. With recovery of hip flexion on the impaired side, abnormal intrajoint coupling and compensatory movement performed by the healthy side will also be alleviated. Furthermore, recovery of spatial symmetry at the peak of hip flexion and extension (SAP) also contributed to improving the symmetry of the range of joint motion, which is reflected by changes in SI$_{ROM}$.

\begin{figure}[tb]
  \centering
  \includegraphics[width=0.465\textwidth]{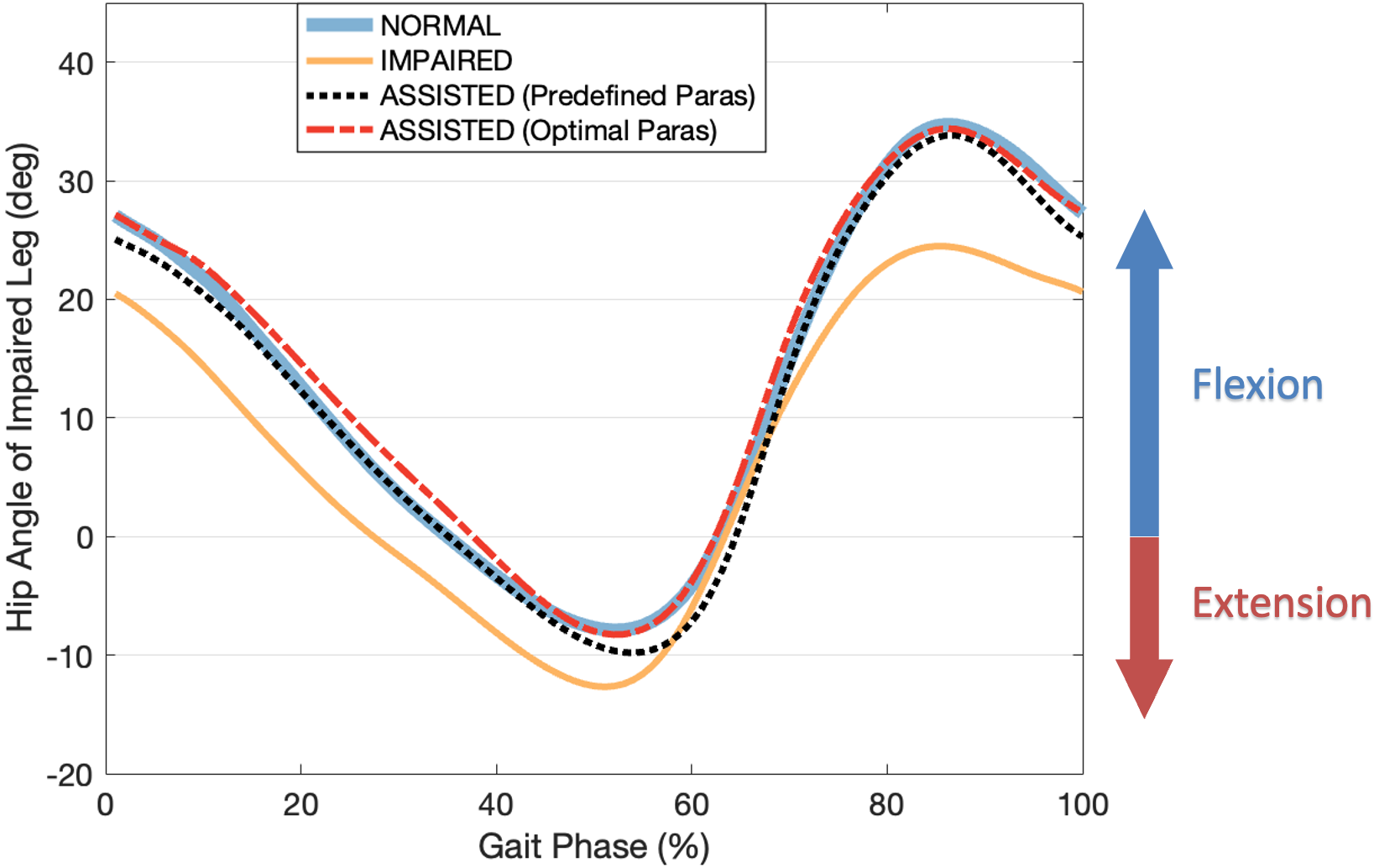}
  \caption{Hip angle time trajectories under different experimental conditions. }
  \label{results_hipangle}
\end{figure}

The ultimate goal of the presented control is to restore normal and symmetric gait throughout the gait cycle for stroke patients, rather than only improve gait asymmetry at landmark gait events such as peak hip flexion and extension. In this context, SI$_{TOA}$ was utilized to evaluate the symmetry of the time trajectory of hip joint angle during continuous walking. As shown in Fig. \ref{results_bar}(d), S$_{TOA}$ decreased from $0.985\, \pm \, 0.003$ (NORMAL) to $0.754\, \pm \, 0.034$ (IMPAIRED) with the artificial gait impairment in place, suggesting significantly worsen gait symmetry during the whole gait cycle. As a contrast, S$_{TOA}$ recovered to $0.957\, \pm \, 0.011$ (ASSISTED with Optimal Paras) with the exoskeleton assistance applied, which is comparable to NORMAL and validates the effectiveness of the proposed control in improving gait symmetry during the whole gait cycle (see Fig. \ref{results_hipangle}).

Apart from the gait symmetry, the subjects' active participation is also crucial to evaluate the effectiveness of the proposed AAN control. As can be seen from Fig. \ref{results_demo}(a), the adjustments of exoskeleton assistance aligned well with changes in the subject's active participation represented by the peak hip flexion moment, which was independently measured with the motion capture system and instrumented treadmill. As the control parameters were updated and the gait symmetry was improved, the exoskeleton assistance plateaued at a reduced and suitable level while the subject's active participation was gradually promoted. As illustrated in Fig.\ref{results_bar}(e), the human participation index, HRI, was improved from $1.442\, \pm \, 0.073$ (IMPAIRED) to $1.612\, \pm \, 0.131$ (ASSISTED with Predefined Paras), and then further improved to $1.774\, \pm \, 0.138$ (ASSISTED with Optimal Paras). The experimental results successfully validated the hypothesis that AAN control with optimal control parameters can stimulate the maximum participation of subjects. Besides, a higher HPI could be observed under the IMPAIRED condition compared to that under the NORMAL condition, which is supposed to be a subconscious response of healthy subjects to the applied artificial resistance for hip flexion.

\section{Discussion and Conclusion}

In this study, an advanced hip exoskeleton control obeying the AAN principle was proposed for gait asymmetry correction of stroke patients, and the key control parameters were online optimized in order to appropriately share control between the exoskeleton and patients. To the best of our knowledge, few AAN exoskeleton controls have been specifically developed to correct the gait asymmetry of stroke patients and successfully address the personalization of key control parameters in such a human-in-the-loop manner.

As defined in (\ref{Objective_AAN}) and demonstrated in Fig. \ref{results_demo}(a), the symmetry errors between the impaired and healthy sides and the magnitude of assistive torque were utilized to evaluate the patient's gait performance and active participation, thereby providing suitable walking assistance accordingly. Compared to most AAN exoskeleton controls\cite{banala2008robot, banala2010novel, srivastava2014assist, duschau2009path, hussain2016assist, aguirre2019phase}, the proposed AAN control stimulated the patients' participation further by not following a predefined joint trajectory and only focused on minimizing the symmetry errors through as little assistance as possible, which allows the patients to choose their own gait patterns and thus makes patient-driven gait rehabilitation possible. However, the symmetry errors are still insufficient for the evaluation of the patient's active participation or motor ability, which is critical for the realization of AAN concept. In this study, given that the interaction force is the only medium for the bidirectional energy exchange between the exoskeleton and patients, the magnitude of assistive torque was utilized to indirectly reflect the patient's participation level during gait rehabilitation. Benefiting from adopting the nSEA-driven hip exoskeleton with accurate sensing and control of the interaction force, no additional expensive force sensor was required. Compared to existing AAN controls that perform evaluations directly based on the motion errors and the interaction force \cite{emken2007human, perez2015assist}, the proposed control took the natural human movement variability into account in the objective function, and thus struck a good balance between improving gait performance and promoting the patient's active participation. By introducing an activation function for the symmetry errors, the proposed control would gradually focus on encouraging the patient's active participation when the symmetry errors fall into the range of acceptable variability.

\begin{table*}
\centering
\caption{Comparison of representative lower-limb exoskeleton controls with human-in-the-loop}
\begin{tabular}{ccccccc} 
\toprule
  Research work  & Device & Optimization Method & \thead{Number of \\Parameters} & Update Frequency & Optimization Time & Optimization Goal\\
  \midrule

   Zhang\cite{zhang2017human}     & Ankle Exo & \thead{Covariance Matrix Adaptation \\ Evolution Strategy (CMA-ES)} & 4 & 16 mins & 64 mins & Reduce energy expenditure \\

   Ding \cite{ding2018human}     & Hip Exo     & Bayesian Optimization (BO)  & 2 & 2 mins & 21.4 $\pm$ 1.0 mins & Reduce energy expenditure  \\

   Zhang \cite{zhang2022imposing}  & Hip Exo & Reinforcement Learning (RL) & 3 & \thead{5 GCs for action\\75 Gcs for policy} & 8.0 $\pm$ 0.7 mins & \thead{Improve hip joint's ROM}  \\

   Ours                          & Hip Exo & Bayesian Optimization (BO)  & 2 & 20 GCs & \thead{5.9 $\pm$ 3.2 mins \\ (312 $\pm$ 120 GCs)} & \thead{Assist as needed for gait \\ asymmetry correction}  \\
  \bottomrule
  \end{tabular}
  \begin{tablenotes}
  \item  GC -- Gait Cycle; ROM -- Range of Motion.
  \end{tablenotes}
  \label{compatison_table}
\end{table*}

To close the human-robot loop on human performance, one of the key challenges is the efficiency of online optimization, especially for individuals with physical impairment and limited endurance \cite{slade2024HIL} (see Table. \ref{compatison_table}). In this study, the online optimization could be finished in an average of 312 $\pm$ 120 gait cycles (or 5.9 $\pm$ 3.2 minutes), and the efficiency was mainly improved from the selection of evaluation metrics and the AAN control scheme. Firstly, the objective function defined in (\ref{Objective_AAN}) only relied on metrics (the symmetry errors and the magnitude of assistive force) that can be evaluated and updated for each gait cycle. In contrast, most assistive exoskeleton controls with human-in-the-loop targeted at reducing the energy expenditure of users\cite{zhang2017human, ding2018human, kantharaju2022reducing}, and thus the objective function required lengthy evaluation times to collect accurate and enough data (metabolic cost). Secondly, the proposed AAN control was developed based on an effective assistive control scheme for gait asymmetry correction. As demonstrated in Fig. \ref{results_bar}, the predefined parameters, albeit not optimal, were still effective. Thus the burden on online optimization was significantly alleviated and only fine adjustments of control parameters were required, resulting in high efficiency of online optimization and great clinical value. Interestingly, similar thinking was also utilized in \cite{li2021toward} to improve the efficiency of online optimization through a combination of offline and online optimizations.

For robot-aided gait rehabilitation obeying the AAN principle, another important question that needs to answer is whether AAN control strategies can actually stimulate the active participation of patients. To the best of our knowledge, few studies have provided direct and solid evidence for this question, while the presented study gives a positive answer from the perspective of biomechanics. As illustrated in Fig.\ref{results_bar}(e), the human participation index (HRI) under the ASSIST (Optimal Paras) condition was significantly improved by 23.0$\%$ compared to that under the IMPAIRED condition. This improvement was independently measured by the motion capture system and instrumented treadmill, ensuring the reliability and accuracy of the results. Besides, surface electromyographic (sEMG) is also a useful approach to evaluate the participation level of patients. But it was not adopted in this study because the attachment of the electrodes to the legs is time-consuming and the reliability of sEMG sensing during long-time gait rehabilitation is still challenging, which might be significantly affected by the skin condition, humidity, and electrode location \cite{riener2005patient, huo2014lower, huo2018fast}.

Although promising results have been yielded, the presented AAN control and proof-of-concept study still have some limitations. The primary limitation is the limited evaluation on healthy subjects with artificial gait impairment, because the focus of this work is on developing advanced AAN control and validating its feasibility and effectiveness in the targeted application setting. Systematic evaluation on the target population, stroke patients with partial motor control, is needed to show its clinical value in the future. Furthermore, in this study, a unilateral hip exoskeleton was adopted to provide hip flexion assistance for the impaired side, which alleviates the intrajoint abnormal coupling and thus indirectly facilitates the recovery of hip extension on the healthy side. However, we noted that the recovery of gait symmetry at peak hip extension was not as fast as that at peak hip flexion, which may relate to the lack of direct control over the healthy side's movement. Thus a bilateral hip exoskeleton with different control strategies for the impaired and healthy sides may be helpful to expedite the process of gait symmetry recovery and further improve gait performance, which can be investigated in our future study.



\bibliographystyle{IEEEtran}
\bibliography{Main}

\end{document}